\documentclass[10pt,twocolumn,letterpaper]{article}

\usepackage{cvpr}              

\usepackage[dvipsnames]{xcolor}

\definecolor{cvprblue}{rgb}{0.21,0.49,0.74}
\usepackage{amsmath,mathtools}
\usepackage{csquotes} 
\usepackage[pagebackref,breaklinks,colorlinks,citecolor=cvprblue]{hyperref}

\def\paperID{26} 
\def\confName{CVPR}
\def\confYear{2024}

\title{Investigating Calibration and Corruption Robustness of Post-hoc Pruned Perception CNNs: An Image Classification Benchmark Study}
\author{
Pallavi Mitra\\
Continental AG, Germany\\
{\tt\small first.last@continental.com}
\and
Gesina Schwalbe\\
University of Lübeck, Germany\\
{\tt\small first.last@uni-luebeck.de}
\and
Nadja Klein\\
TU Dortmund, Germany\\
{\tt\small first.last@tu-dortmund.de}
}

\renewcommand*{\paragraph}[1]{\noindent\textbf{#1}~}

\begin{document}

\maketitle
\raggedbottom
\begin{abstract}
Convolutional Neural Networks (CNNs) have achieved state-of-the-art performance in many computer vision tasks. However, high computational and storage demands hinder their deployment into resource-constrained environments, such as embedded devices. Model pruning helps to meet these restrictions by reducing the model size, while maintaining superior performance.
Meanwhile, safety-critical applications pose more than just resource and performance constraints. In particular, predictions must not be overly confident, i.e., provide properly calibrated uncertainty estimations (proper \emph{uncertainty calibration}), and CNNs must be robust against corruptions like naturally occurring input perturbations (\emph{natural corruption robustness}). This work investigates the important trade-off between uncertainty calibration, natural corruption robustness, and performance for current state-of-research post-hoc CNN pruning techniques in the context of image classification tasks. Our study reveals that post-hoc pruning substantially improves the model's uncertainty calibration, performance, and natural corruption robustness, sparking hope for safe and robust embedded CNNs. 
Furthermore, uncertainty calibration and natural corruption robustness are not mutually exclusive targets under pruning, as evidenced by the improved safety aspects obtained by post-hoc unstructured pruning with increasing compression.
\end{abstract}    
\section{Introduction}
\label{sec:intro}

In the realm of computer vision, Convolutional Neural Networks (CNNs) have emerged as a dominant paradigm, demonstrating remarkable success in diverse applications, including image classification \cite{yang2020cars}, object detection \cite{zhao2019object} and video analysis \cite{villegas2019high}. 
However, due to their extensive parameter count \cite{denton2014exploiting}, these architectures demand substantial storage and computational resources, posing limitations on deployment in embedded systems. To address this issue, pruning methods have been introduced that reduce the number of parameters, effectively compress the network, and decrease computational complexity  \cite{molchanov2019importance, ghosh2023pruning}. 
The primary strategy involves identifying and eliminating the least important network components while preserving desired performance measures, such as accuracy. The 
pruning landscape consists of two main categories \cite{liang2021pruning}: unstructured and structured pruning. Unstructured pruning deals with the individual weights of a network, resulting in the development of sparse models. In contrast, structured pruning typically eliminates complete channels, filters, or layers. Additionally, concerning the timing of pruning during the training of a neural network, pruning algorithms can be categorized into two groups: post-hoc and ante-hoc pruning \cite{yue2019really}. Post-hoc pruning algorithms \cite{han2015learning, li2016pruning, liu2017learning} operate on pre-trained models, utilizing knowledge from the initial training to selectively remove parameters based on criteria. In contrast, ante-hoc pruning algorithms \cite{lee2018snip, frankle2018lottery} explore effective model architectures during the pre-training phase. Of particular interest for this paper is post-hoc pruning, 
i.e., pruning of an already trained CNN. In contrast to ante-hoc pruning algorithms that need intervention during training, this allows the distribution of training tasks and integration to different teams or suppliers. Irrespective of the diverse pruning methodologies, pruning algorithms primarily focus on assessing post-pruning accuracy and inference time. Nevertheless, these metrics alone do not provide a comprehensive understanding of other consequences of pruning, such as the impact on model robustness and reliability \cite{goldstein2020reliability,kaur2022deadwooding}.

In real-world applications, CNN architectures must be robust against perturbations of the inputs \cite{schwalbe2020structuring,willers2020safety}, such as out-of-distribution data \cite{yang2022generalized}. As of the categorization in \cite{schwalbe2020structuring, willers2020safety}, common robustness challenges include adversarial attacks, concept drift, and covariate shift. In the case of adversarial attacks, intentional manipulations aim to deceive the model \cite{meyers2023safetycritical}. Concept drift refers to a task-specific output label distribution change due to, e.g., novel objects \cite{widmer1996learning}. Hence, for assessments of architecture-specific robustness, task-agnostic covariate shifts are most relevant, which means changes in the input data distribution over time due to perturbations such as sensor noise, varying weather conditions, and so on \cite{amodei2016concrete}.

Several studies demonstrate the vulnerability of CNNs to typical data distortions, compared to human performance \cite{geirhos2018generalisation}. For example, in assessments of image recognition models, CNNs were found to be notably susceptible to covariant shift such as blurring and Gaussian noise \cite{dodge2016understanding}. The absence of invariance to slight translations in multiple CNNs was also identified by \cite{azulay2018deep}. Even worse, when it comes to real-world conditions, Von Bernuth et al. \cite{von2019simulating} report a significant performance degradation of an object detection CNN when evaluated on weather-corrupted data. Meanwhile, state-of-the-art methods for enhancing the robustness of CNNs typically rely on large models \cite{hendrycks2018benchmarking}, impractical for resource-constrained settings. Pruning resolves this issue by reducing memory and computational demands. There has been extensive research on the robustness of pruned models against adversarial attacks \cite{sehwag2020hydra, goldblum2020adversarially, gui2019model}; however, adversarial robustness does not necessarily extend to other corruptions like covariate shifts. Therefore, the trade-offs related to natural corruption robustness in pruned networks remain undetermined.

Moreover, while deep learning research focuses on improving the accuracy of networks, less attention has been given to the reliability of the networks. Model uncertainty calibration represents the degree of correspondence between the model output confidences and their actual probability of correctness \cite{minderer2021revisiting}, ensures well-calibrated confidences as a reliable indicator of output trustworthiness. In recent years, a growing body of research has highlighted a respective trend: despite accuracy advancements, modern neural networks often exhibit poor uncertainty calibration \cite{guo2017calibration}. However, in real-world safety-relevant applications, classification networks must be both accurate and well-calibrated to trigger appropriate safety measures during high uncertainty \cite{bojarski2016end,willers2020safety}.
Pruning introduces structural changes and weight reductions in neural networks, potentially affecting their predictive capabilities and confidence estimates \cite{yuan2022membership}. Despite this, the impact of different post-hoc pruning approaches on uncertainty calibration, including the trade-off between compression and calibration, enjoyed little attention so far.

Since state-of-the-art post-hoc pruning techniques purely consider CNN accuracy for guiding the compression, it is not obvious that pruning will preserve any other safety targets; the structural changes might even result in degradation thereof, rendering pruning inapplicable to safety-relevant applications. For this reason, in this paper, we systematically investigate the uncertainty calibration error and natural corruption robustness of post-hoc pruned CNNs under increasing compression compared to the original network for image classification task. To the best of our knowledge, this is the first research work in the literature benchmarking several popular unstructured and structured post-hoc pruning techniques for image classification while thoroughly and quantitatively studying their uncertainty calibration and natural corruption robustness trade-offs. It provides a detailed understanding of the foremost positive influence of post-hoc pruning on safety properties. Our key findings are:

\begin{enumerate}
    \item Post-hoc pruning consistently improves the uncertainty calibration compared to their unpruned counterparts; post-hoc unstructured pruned models exhibit substantial improvements in calibration, whereas post-hoc structured pruned models exhibit improved uncertainty calibration performance up to a specific degree of compression.
    \item Post-hoc pruning has no negative impact on natural corruption robustness.
    \item Post-hoc pruning does not affect uncertainty calibration in the presence of natural corruption.
\end{enumerate}

\section{Related Work} \label{sec:related work}

\subsection{Safety Metrics} \label{sec:related work.safety metrics}

When assessing CNNs, most works concentrate on standard performance metrics such as accuracy and F1 score (for classification) or mean average precision (for object detection). To ensure the safety of CNNs, one needs to consider their technology-specific insufficiencies. Sämman et al. \cite{samann2020strategy} and Schwalbe et al. \cite{schwalbe2020structuring} categorized them into:
lack of generalized \emph{performance};
incorrect internal \emph{logic};
lack of \emph{robustness} against perturbations that do not change the semantic content of the input;
lack of \emph{efficiency} for the respective hardware, usually correlated with the CNN size; and
CNN \emph{opaqueness}, which, however, is hard to quantify across use-cases objectively \cite{zhou2021evaluating,schwalbe2023comprehensive}.  
This catalogue was extended to consider incorrect \emph{uncertainty} estimate outputs separately 
measured by variants of the calibration error metric - expected, average, and maximum calibration error \cite{guo2017calibration}.
Some safety metrics found in the literature are specifically tailored to the safety needs as well as potential logical issues tied to a use-case \cite{willers2020safety,schwalbe2020survey}. Examples are 
consistency with logical constraints 
\cite{schwalbe2022enabling} or standard detection accuracy weighted by safety-relevance of objects \cite{cheng2018dependability}. However, such metrics require a concrete reference system and are hardly generalized across use cases.
Robustness against perturbations, however, gives rise to a rich set of application-agnostic metrics. As of \cite{schwalbe2020structuring} and \cite{willers2020safety}, these are divided into two classes: adversarial and natural corruption robustness. The first considers robustness against adversarial attacks, i.e., targeted and maliciously crafted changes to the input \cite{carlini2017evaluating,wei2023visually,meyers2023safetycritical}, and sets focus on security. On the other side, natural corruption robustness considers naturally occurring non-semantic corruptions of inputs, such as noise and translations from sensor degradation, adverse weather conditions like fog and rain \cite{schwalbe2020structuring}. Percentage change in accuracy, the relative change in failure rate, mean performance under corruption, and mean corruption error are the most popular metrics for both robustness against adversarial attacks and natural corruption robustness \cite{meyers2023safetycritical, hendrycks2018benchmarking}. 
Our work focuses on \emph{expected calibration error} and \emph{mean performance under corruption} as the safety metrics for uncertainty calibration and natural corruption robustness, respectively. The aforementioned safety measures are explored mostly for vanilla networks without considering model compression.

\subsection{Model Compression}\label{sec:related work.compression}
In order to leverage the power of large state-of-the-art computer vision CNNs in resource-constrained environments, compression techniques must be employed to reduce the number of parameters and, thus, computation operations and memory of the deployed CNN.
As of He et al. \cite{he2018amc}, typical and complementary means of compression are quantization, i.e., compression of the weight value representation \cite{rokh2023comprehensive}, and CNN pruning. Pruning started as early as the 1980s \cite{reed1993pruning}, and, unlike hardware-dependent quantization, aims to remove parameters on an architectural level.
Following \cite{liang2021pruning}; pruning methods can be differentiated by their application time (dynamic during operation versus static before deployment), their pruning criterion, the removed elements, and whether they are structured or unstructured.
Operation-time pruning employs decision logic at runtime to dynamically remove computational paths \cite{liang2021pruning}, but cannot decrease CNN memory consumption.

{Static pruning approaches can be sub-classified according to their instance of interference before deployment:
1. \emph{Post-hoc: Pruning after training} is a three-step process: first, train the initial network to convergence; second, prune redundant parameters based on specific criteria; and finally, retrain the pruned model (fine-tuning) to recover any performance loss incurred during the pruning process \cite{han2015deep, li2016pruning, liu2017learning}.
2. \emph{Ante-hoc:}
 (a)\emph{Pruning during training:} compared to pruning after training, connections are dynamically deactivated during training based on their importance. But later weights can adapt and potentially be reactivated based on gradient updates \cite{zhu2017prune}. 
 (b)\emph{Pruning before training resp.~weight rewinding:} motivated by the Lottery Ticket Hypothesis \cite{frankle2018lottery} some recent studies aim to identify a sparsity mask that can be used for weight initialization, e.g., using information from a previous training run, and subsequently train the pruned network from scratch while maintaining the original mask throughout the training process \cite{lee2018snip, wang2020picking}.
In this article, we have adopted approach 1., i.e., pruning after training or post-hoc pruning, which {not only} has the largest body of research \cite{le2021network}{, but also most practical relevance due to the separation of concerns of model training and integration}. 
In post-hoc pruning, the model can be pruned with different compression ratios without the necessity of initiating training from its initial state. This affords flexibility in modifying compression ratios without undertaking a comprehensive model retraining. Conversely, when employing ante-hoc algorithms for model pruning with diverse compression ratios, it is imperative to initiate model retraining to accommodate the specific compression ratio, contrasting the more adaptable approach offered by post-hoc pruning.}

In our work, we consider two fundamental categories of static post-hoc pruning: unstructured pruning and structured pruning. \emph{Unstructured pruning} involves the removal of individual weights which are assigned the least importance for the network functioning, resulting in a sparse network without sacrificing predictive performance \cite{han2015learning, han2015deep}. However, since the positions of non-zero weights are irregular and random, the sparse network pruned by unstructured pruning cannot be presented in a \emph{structured} fashion. This means that it cannot lead to compression and speedup without dedicated hardware or libraries \cite{han2016eie}. To address this, so-called \emph{structured pruning} typically prunes complete channels, filters, or layers within the CNN architectural structure according to an importance criterion \cite{li2016pruning, he2017channel, liu2017learning}. As a result, the pruned network retains the original convolutional structure without introducing sparsity, and no sparse libraries or specialized hardware are required to realize the benefits of pruning.
Since both have practical advantages, we cover and compare these two pruning paradigms in this study. 

For ease of notation, pruning refers to post-hoc pruning in the remainder of this paper.

\subsection{Pruning and Uncertainty Calibration}
There is a limited exploration of the impact of pruning on uncertainty calibration. 
While a study conducted by Sun et al. \cite{sun2021sparse} indicates that sparsification is beneficial for enhancing the calibration of CNNs, their study specifically focused on various unstructured pruning methods applied to smaller residual networks (ResNet-20, ResNet-32) with lower sparsity levels (up to 20\%). Consequently, these investigations lack a comprehensive analysis of diverse pruning methodologies involving larger residual networks and higher sparsity levels. This gap in research underscores the fact that the impact of pruning on uncertainty calibration remains both active and largely unexplored which we aim to tackle here.

\subsection{Pruning and Robustness}
There is a growing body of literature on the robustness of pruning methods against adversarial attacks, focusing mainly on the improvement of the adversarial robustness of pruned models even without adversarial training \cite{jordao2021effect, guo2018sparse}. According to these results, pruned models typically do \emph{not} inherit the susceptibility to adversarial attacks observed in the original models. 
The previously cited studies demonstrate the positive effect of pruning concerning robustness against adversarial attacks. However, this is limited to adversarial robustness, \emph{not taking into account} robustness against naturally occurring corruptions like fog or random noise from sensor degradation. 

Compared to research on adversarial robustness, there is a limited exploration of the impact of pruning on natural corruption robustness. Regarding the relationship between pruning and natural corruption robustness, the study conducted by Hooker et al. \cite{hooker2019compressed} revealed that the corruption performance significantly deteriorates when higher pruning ratios are applied for the magnitude pruning method.
According to Hoffmann et al. \cite{hoffmann2021towards}, unstructured (\emph{global weight}) pruning preserves more robustness regarding performance under corruption than the structured (\emph{$L_{1}$-norm filter}) pruning counterparts. However, such studies lack a comprehensive examination of diverse pruning methodologies, leaving the influence of pruning on natural corruption robustness an active and unexplored research area.
\section{Methods} \label{sec:method & metric}
\subsection{Pruning} \label{sec:method & metric.pruning}
All static pruning methods commonly consist of a \emph{step 1}, in which they determine an importance score for candidate network units, \emph{step 2}, in which this is used to select and remove items with a low score, and an optional last step of fine-tuning the newly obtained pruned network.
The decision threshold in the second step influences the final \emph{pruning ratio}, i.e., the percentage of removed parameters, which can serve as a measure for the achieved compression.

As outlined in Section\,\ref{sec:related work.compression}, static pruning approaches that apply after model training can be divided into \emph{unstructured pruning} of weights, and \emph{structured pruning} of filters or channels in case of CNNs \cite{liang2021pruning}.
To evaluate pruning with respect to uncertainty calibration and natural corruption robustness and to compare different pruning paradigms with respect to these aspects, we consider the following three pruning methods, which are commonly used and chosen to cover a broad range of static pruning approaches \cite{le2020network}. 

\subsubsection{Unstructured Weight Pruning} \label{WP}
We consider the unstructured pruning strategy from Han et al. \cite{han2015learning} for magnitude-based weight pruning, comprising two steps. \emph{Step~1:} globally
sort the weights according to their relative importance based on the magnitude of the weights calculated by $L_{1}$-norm. \emph{Step~2:} prune ${k}\%$ of the weights with the lowest importance, where $k\%$ is the pruning ratio.

\subsubsection{Structured Filter Pruning} \label{FP}
We adopt the $L_{1}$-norm-based filter pruning technique by Li et al. \cite{li2016pruning} as a filter-level structured pruning method. \emph{Step~1:} rank the filters by their $L_{1}$-norm value in each convolutional layer. \emph{Step~2:}  remove the $k\%$ lowest ranking filters, where $k\%$ approximates the pruning ratio.

\subsubsection{Structured Channel Pruning} \label{CP}
Additionally, we employ network slimming introduced by Liu et al. \cite{liu2017learning} as another structured pruning method targeting channel-level sparsity.
\emph{Step~1:} During training, impose $L_{1}$ sparsity regularization on the channel-wise scaling factors for each channel from batch normalization layers.
\emph{Step~2:} Remove those channels with near-zero scaling factors afterwards. 

\subsection{Safety Metrics} \label{sec:method & metric.metrics}
Besides efficiency, the two main application-agnostic safety targets for computer vision CNNs are correct uncertainty calibration and robustness against naturally occurring corruptions \cite{schwalbe2020structuring,meyers2023safetycritical,zhou2021evaluating}.
These are quantified by means of expected calibration error and mean performance under corruption on a benchmark corruption dataset, respectively. 

\subsubsection{Uncertainty Calibration} \label{ECE}
In the context of employing pruned models in safety-critical applications, ensuring good uncertainty calibration is of high importance, in addition to achieving the desired levels of sharpness of uncertainty. 
In this paper, we want to investigate whether pruning has an adverse effect on CNN uncertainty calibration. Therefore, the calibration error concept is employed to re-assess the uncertainty calibration properties of the pruned models with respect to the original unpruned model.

Good uncertainty calibration aims to ensure that the predicted confidences correctly represent the actual probability of the correctness of the prediction. Miscalibration is commonly quantified in terms of Expected Calibration Error (ECE) \cite{guo2017calibration}. This is measured by first binning samples of the test set according to their predicted confidence value; then measuring the accuracy for each bin; and lastly, determining the weighted average of the difference between bins’ accuracy and mean predicted confidence. This formalizes to 
\begin{gather}
\SwapAboveDisplaySkip
\mathit{ECE} = \sum_{m=1}^{M}\frac{B_{m}}{n}\left | \mathit{acc}(B_{m}) - \mathit{conf}(B_{m}) \right |,
\label{eq:ete}
\end{gather}
where $M$ is the total number of bins into which the predictions are equally grouped, $B_{m}$ is the number of samples whose prediction confidence falls into
the interval $I_{m} = \left(\frac{m-1}{M}, \frac{m}{M}\right]$ and $n$ is the total number of samples. 
The accuracy of $B_{m}$ is defined as:
\begin{align}
    \SwapAboveDisplaySkip
    \mathit{acc}({B_{m}}) = \frac{1}{\left | B_{m} \right |}\sum_{i\in  B_{m}}\mathcal{I}(\hat{y}_{i} = {y_{i}}),
\end{align}
where $\hat{y}_{i}$ and ${y_{i}}$ are the predicted and true class labels for
sample $i$, and $\mathcal{I}(\cdot)$ is the indicator function. 
Finally, the average confidence within bin $B_{m}$ is defined as:
\begin{align}
    \mathit{conf}({B_{m}}) = \frac{1}{\left | B_{m} \right |}\sum_{i\in  B_{m}  }\hat{p}_{i},
\end{align}
where $\hat{p}_{i}$ is the confidence for sample  $i$.

\subsubsection{Natural Corruption Robustness} \label{mPC}
Robustness to distribution shift is an important feature of CNNs for real-world applications, where the environmental conditions may vary substantially. Among many forms of distribution shift, one particularly relevant category for computer vision is covariate shift, i.e., input image corruption \cite{amodei2016concrete,willers2020safety,schwalbe2020structuring}. Therefore, natural corruption robustness is important when deploying pruned models in safety-critical applications. This paper investigates how pruning influences natural corruption robustness compared to the original unpruned model. This inquiry aims to provide insights into the potential trade-offs and implications of pruning methodologies concerning the network's ability to withstand natural corruption in safety-critical applications.

The natural corruption robustness of a (pruned or unpruned) model is evaluated in terms of \emph{mean performance under corruption}($\mathit{mPC}$) \cite{michaelis2019benchmarking}, which is defined as
\begin{equation}\label{eq:mPC}
\mathit{mPC} = \frac{1}{N_{c}}\sum_{c=1}^{N_{c}}\frac{1}{N_{s}}\sum_{s=1}^{N_{s}}P_{c,s},
\end{equation}
where $P_{c,s}$ represents the performance computed on test data corrupted with corruption type $c$ under severity levels $s$. $N_{s}$ and $N_{c}$ denote the number of severity levels and corruptions respectively. 

\section{Experimental Setup} \label{sec:experimental setup}

The experimental setup for benchmarking the pruning methods from Section\,\ref{sec:method & metric.pruning} against the safety metrics from Section\,\ref{sec:method & metric.metrics} and accuracy is detailed below.

Concretely, the investigated research questions are: 

\begin{enumerate}[label=(\roman*)]
    \item Do increasing pruning ratios affect any of
        \begin{enumerate}[label=\alph*.]
            \item uncertainty calibration (see Section\,\ref{sec:result_1});
            \item natural corruption robustness (see Section\,\ref{sec:result_2}); or
            \item uncertainty calibration when challenged with natural corruptions (see Section\,\ref{sec:result_3});
        \end{enumerate}
        
    \item  Is there a difference between structured and unstructured pruning in any of the above cases?
\end{enumerate}
    

\subsection{Datasets \& Models} \label{sec:data}

Within the network pruning literature, CIFAR-10 stands as the established benchmark dataset, and VGG, ResNet serve as the prevalent network architectures. Our assessment of the three pruning methods aligns with the target models and dataset pairs presented in the original paper, ensuring the comparability of our results.

\paragraph{CIFAR-10:} A common benchmark dataset for image classification is CIFAR-10~\cite{krizhevsky2009learning}. It contains
60,000 (50,000 training and 10,000 test images) colour images of $32\times 32$ resolution in 10 classes, with an equal distribution of 6,000 images per class. We use the CIFAR-10 training data as in-distribution data (clean data) to train the CNNs and to fine-tune the pruned models. The test split is used to determine the safety metrics of unpruned and pruned models on clean data.

\paragraph{CIFAR-10-C:} This dataset is constructed by synthetically corrupting the original CIFAR test sets \cite{hendrycks2019benchmarking}. It consists of 15 types of corruption, each further categorized into five distinct severity levels, containing 50,000 images for each type of corruption. The corruptions cover four categories: noise, blur, weather effects, and digital transforms.
In this paper, CIFAR-10-C is used as naturally corrupted data during testing to check the natural corruption robustness of original and pruned CNN models \cite{minderer2021revisiting}.

\paragraph{Models:}
VGG networks, introduced by Simonyan and Zisserman \cite{simonyan2014very}, leverage a deep architecture with small convolutional filters, showcasing robust performance in image classification task \cite{alzubaidi2021review}. 
Residual network backbones, pioneered by He et al. \cite{he2016deep}, prove highly effective in mitigating the vanishing gradient problem, enabling the training of extremely deep networks and achieving state-of-the-art results in image classification tasks \cite{alzubaidi2021review}. 

To ensure comparability between unstructured and structured pruning, we choose VGG-19 and 
PreResNet-110 for weight pruning, VGG-16 and 
ResNet-110 for filter pruning and VGG-19 and 
ResNet-164 for the more rigorous channel pruning. This setup is able to reproduce accuracy comparable to the reported results of the baseline models from the original works (for accuracy on clean test data, see Figure~\ref{fig:Calibration error}). 

\subsection{Training Configuration}
We adopt the implementation and hyperparameters for weight pruning, filter pruning, and channel pruning from the publicly available codebase by Liu et al. \cite{liu2018rethinking}, demonstrating comparable results to the original works. Using a stochastic gradient descent optimizer, the original models are trained for 160 epochs with a batch size of 64. An exponentially decreasing learning rate is applied, starting at 0.1 for epochs [1, 80), 0.01 for epochs [80, 120), and finally 0.001 until epoch 160. 
Simple data augmentation involving random crop and random horizontal flip is used on training images as a standard means to foster natural corruption robustness in the original models. For fine-tuning the pruned models, we use a constant learning rate set to the last one used for training the original model (0.001) and apply this for 40 epochs.

\subsection{Pruning Ratio Selection} \label{sec:experimental setup.pruning ratio}
The pruning ratios vary among the selected pruning methods, each characterized as follows: 

\paragraph{Magnitude-based weight pruning:} The pruning ratio, defined as the percentage of parameters pruned within the convolutional weights of convolution layers, establishes the pruning threshold. It determines which weights are set to zero based on their magnitudes relative to the specified threshold value.

\paragraph{Filter Pruning:} VGG-16 on CIFAR-10 comprises 13 convolutional layers and 2 fully connected layers. Pruning 512-feature map layers, as reported in \cite{li2016pruning}, maintain accuracy due to filters' limited spatial connections on small feature maps. Despite extensive pruning in the first layer, the remaining filters outnumber input channels. However, excessive pruning in the second layer risks losing vital information. To maintain compatibility with the original implementation, our approach selectively prunes layers 1 and 8 to 13.

ResNets designed for CIFAR-10 comprise three stages of residual blocks, handling feature maps of sizes $32\times 32$,
$16\times 16$, and $8\times 8$. Each stage maintains an identical number of residual blocks. Deeper layers exhibit increased sensitivity to pruning compared to earlier stages, as noted in \cite{li2016pruning}. Specifically targeting the first layer of the residual block, our approach ensures compatibility with the original implementation by exclusively pruning the first layer of each residual block within the first stage of the network. The pruning ratio reflects the percentage of filters pruned for all first layers in the first stage.

\paragraph{Channel Pruning:} 
We adopt the approach outlined in \cite{liu2017learning} by utilizing a universal pruning threshold applied consistently across all layers. This threshold is established based on a percentile value among all scaling factors. For instance, we prune channels by selecting those with lower scaling factors, which is achieved by setting the percentile threshold accordingly.

\subsection{Evaluation Metrics}
\paragraph{Performance:}%
We use classification accuracy to measure the performance of original and pruned models on the clean dataset. 

\paragraph{Natural Corruption Robustness:}%
We report mean accuracy under corruption as a performance metric (mPC, see Section\,\ref{mPC}) overall corruption types for each severity level. 

\paragraph{Uncertainty Calibration:}%
To estimate the miscalibration of the original and pruned model on the clean and corrupted datasets, we use $\mathit{ECE}$ (see Section\,\ref{ECE}) using equal-mass binning with ten bins.

\paragraph{Compression:}%
To examine the impact of pruning on the other safety targets, we benchmark the three different pruning methods from Section\,\ref{sec:method & metric.pruning}. The resulting compression is measured in terms of their respective pruning ratio (Section\,\ref{sec:experimental setup.pruning ratio}), which is sampled at a rate of 10\% from values 0-70\%, where 0\% indicates the original network (unpruned).   
\definecolor{matplotlibblue}{HTML}{1f77b4}
\definecolor{matplotliborange}{HTML}{ff7f0e}
\definecolor{matplotlibgreen}{HTML}{2ca02c}

\section{Results} \label{sec:results}
\subsection{Does pruning affect uncertainty calibration?} \label{sec:result_1}

To examine the impact of pruning on network uncertainty calibration, we compared  $\mathit{ECE}$ and accuracy under different pruning ratios for the three selected pruning techniques. The results are shown in  Figure~\ref{fig:Calibration error} for weight, filter, and channel pruning, respectively (we conduct each experiment three times and report $\text{mean}\pm\text{std}$).

The results suggest that \textit{even high pruning ratios do not impact the uncertainty calibration} compared to that of the original unpruned model. In unstructured (weight) pruning, the calibration error for all pruning ratios is less than the calibration error of the original unpruned model. Hence, unstructured pruning can even enhance the uncertainty calibration of the VGG-19 and
ResNet-110 model. Whereas, in filter and channel pruning, the calibration error for pruned models up to a certain pruning ratio (filter pruning: 50\% on VGG-16 and 50\% on ResNet-110, channel pruning: 40\% on VGG-19 and 50\% on ResNet-164) is less than or similar to the calibration error of the original unpruned model. After that, the calibration errors of pruned models do not increase substantially with respect to the calibration error of the unpruned model.

\subsection{Does pruning affect natural corruption robustness?}\label{sec:result_2}

\begin{figure*}
  \centering
  \begin{minipage}{\textwidth}
    \includegraphics[width=\linewidth]{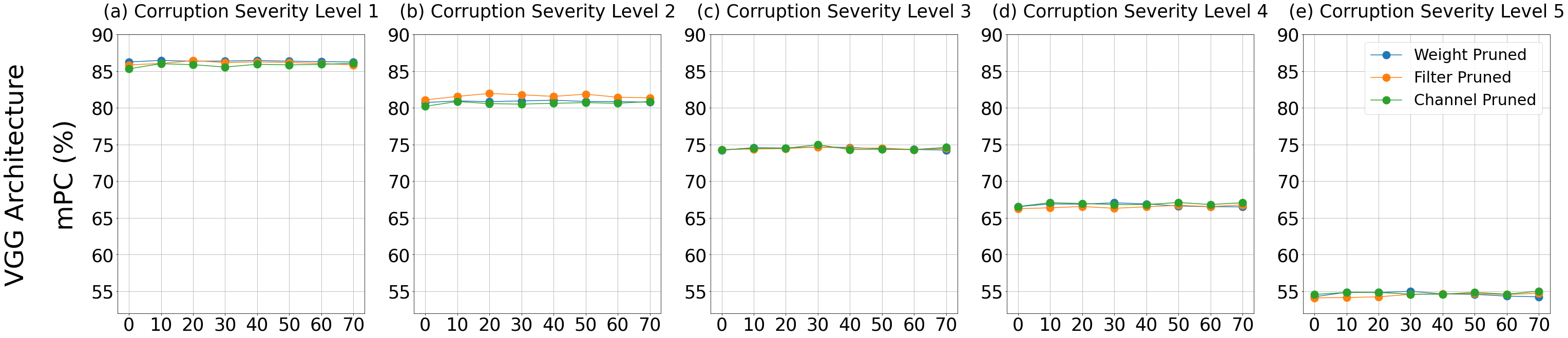}
  \end{minipage}\hfill
  \begin{minipage}{\textwidth}
    \includegraphics[width=\linewidth]{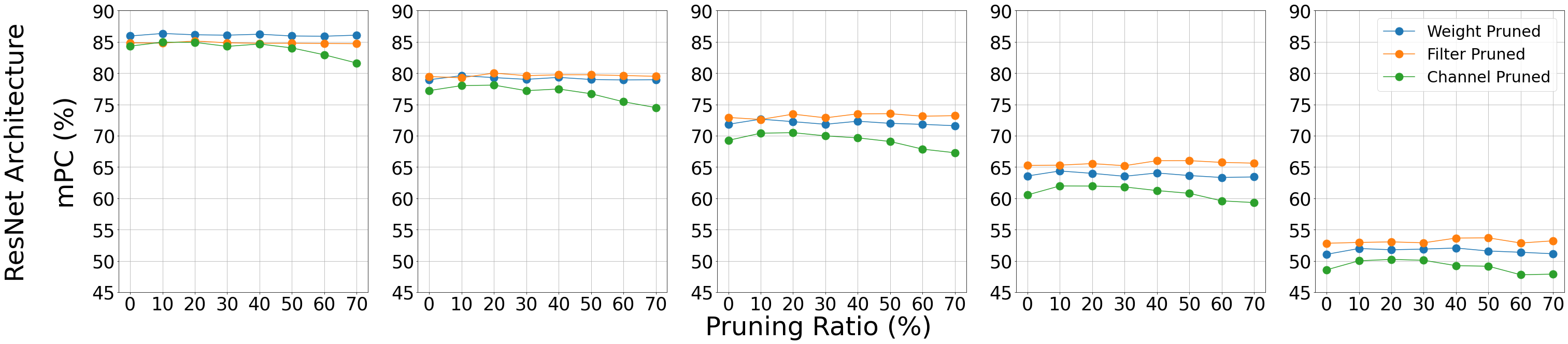}
  \end{minipage}
  \caption{Development of $\mathit{mPC}$\,($\uparrow$) for increasing
pruning ratios for five severity levels of corruption}
\label{fig:Robustness}
\end{figure*}

To answer this question, the choice of pruning methods and ratio are kept similar, as mentioned in Section\,\ref{sec:result_1}, but naturally corrupted data is considered as test data. Here, $\mathit{mPC}$ is measured for unpruned and pruned models for the selected pruning methods from Section\,\ref{sec:method & metric.pruning}. $\mathit{mPC}$ is calculated separately for each of the five corruption severity levels from CIFAR-10-C, each pruning ratio step, and each pruning method. Figure~\ref{fig:Robustness} illustrates how natural corruption robustness is influenced by pruning for different severity levels for weight, filter, and channel pruning, respectively. 
\begin{figure*}
  \centering
  \begin{minipage}{\textwidth}
    \includegraphics[width=\linewidth ]{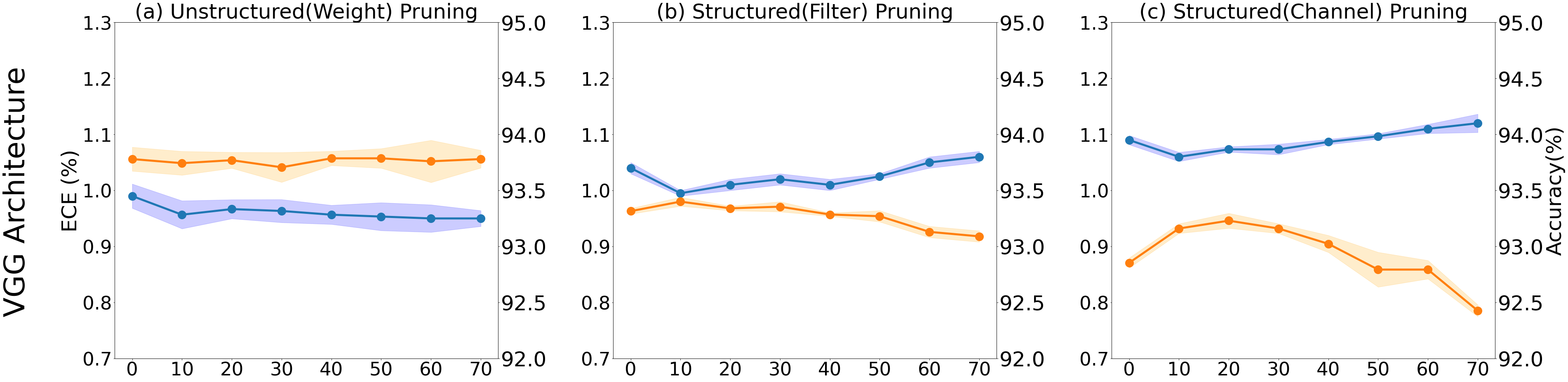}
  \end{minipage}\hfill
  \begin{minipage}{\textwidth}
    \includegraphics[width=\linewidth]{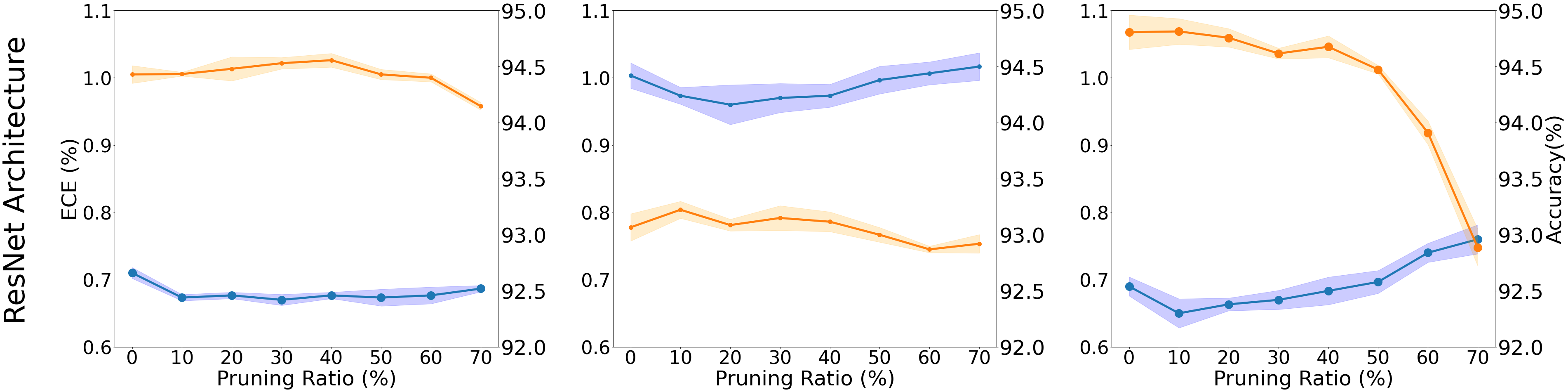}
  \end{minipage}
  \caption{Comparing development of \textcolor{matplotlibblue}{$\mathit{ECE}$}\,($\downarrow$) (\textcolor{matplotlibblue}{blue} lines, left y-axis scaling) and \textcolor{matplotliborange}{accuracy}\,($\uparrow$) (\textcolor{matplotliborange}{orange} lines, right y-axis scaling) under increasing pruning ratios}
\label{fig:Calibration error}
\end{figure*}

\begin{figure*}
  \centering
  \begin{minipage}{\textwidth}
    \includegraphics[width=\linewidth]{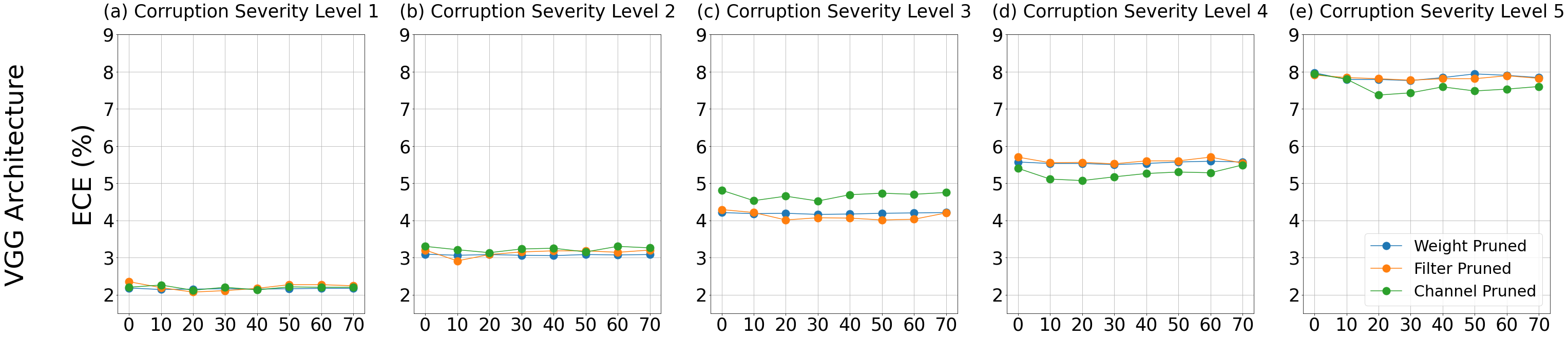}
  \end{minipage}\hfill
  \begin{minipage}{\textwidth}
    \includegraphics[width=\linewidth]{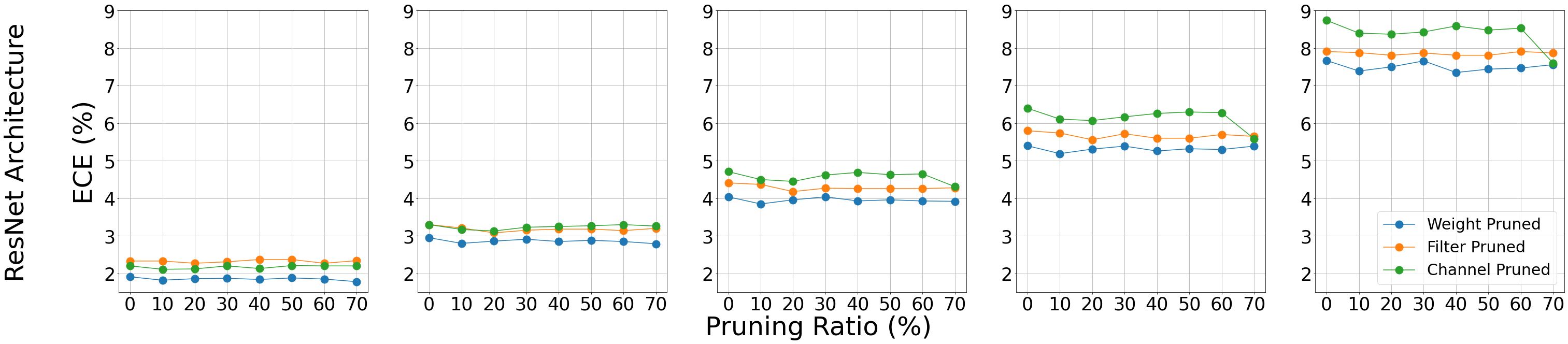}
  \end{minipage}
  \caption{Development of $\mathit{ECE}$\,($\downarrow$) for increasing pruning ratio for five severity levels of corruption} 
 \label{fig:calib_corr}
\end{figure*}

The observation implies that the \textit{natural corruption robustness of weight pruned and filter pruned models, as assessed through $\mathit{mPC}$, is better or similar compared to the original unpruned model}. Notably, the robustness against natural input corruption remains unaffected by the weight pruning of VGG-19, ResNet-110, filter pruning of VGG-16, ResNet-110, and channel pruning of VGG-19  across all severity levels. In contrast, for channel pruning, the $\mathit{mPC}$ of ResNet-164 starts to degrade from ca.~60\% pruning ratio for all corruption levels. Consequently, the robustness experiences deterioration with higher pruning rates for channel pruning in the ResNet-164 model across all severity levels. This suggests that information needed to compensate for corruption is highly distributed over channels, i.e., wider networks (in terms of the number of channels) might have better chances of achieving zero-shot natural corruption robustness than narrow ones.

While the pruning ratio seems to play a negligible role in accuracy in the presence of corruption, one should, however, note the severe drop in accuracy for increasing corruption severity levels (from more than 90\% without corruption in Figure\,\ref{fig:Calibration error} down to less than 55\%). This attests to the generally weak overall robustness of CNNs against strong natural corruption.

\subsection{Does pruning impact uncertainty calibration in the presence of natural corruption?} \label{sec:result_3}
Here, $\mathit{ECE}$ is measured for unpruned and pruned models using the different pruning methods from above in the presence of natural corruption of different severity levels. 
Figure~\ref{fig:calib_corr} illustrates how the uncertainty calibration is influenced by pruning in the presence of natural corruption for weight, filter, and channel pruning, respectively. 

The results demonstrate that \textit{pruning does not negatively impact the model uncertainty calibration compared to the original unpruned model, even when additionally confronted with natural corruptions}. In weight, filter, and channel pruning, the measured calibration error for all pruned models at different pruning ratios is similar to or less than the calibration error of the original unpruned VGG-19, VGG-16, ResNet-110 and ResNet-164 models for all five severity levels of corruption. 
Nevertheless, as for $\mathit{mPC}$, $\mathit{ECE}$ increases rapidly with increasing severity levels and  worse $\mathit{mPC}$,  reaching more than 400\% increase of $\mathit{ECE}$ for the highest level. Hence, natural corruptions seem not only to pose a challenge to robust accuracy but also to trustworthiness in terms of uncertainty calibration.
\section{Conclusion} \label{sec:conclusion}
For safety-critical computer vision applications, model efficiency, proper uncertainty calibration, and natural corruption robustness are---and will be---the key desirables.
This work, for the first time, investigated whether popular post-hoc pruning as a means for model compression conflicts with the other two safety targets. Our benchmark with a standard setup of model, dataset, and post-hoc pruning methods provided promising insights into this: we could \emph{not} find a negative effect of pruning on natural corruption robustness and uncertainty calibration; calibration was not even affected by pruning when challenged with naturally corrupted inputs. Our considered post-hoc unstructured pruning method showed a consistently positive effect on uncertainty calibration even when pruning up to 70\%. 
While our results on typical image classification backends do not yet cover the whole spectrum of computer vision tasks and architectures, they raise hope that accuracy-driven pruning does not contradict but even enhances other safety targets. Future work includes extending our investigation to costly experimental setup tasks like object detection and semantic segmentation.

Further research in the realm of safe pruning could explore tailored safety objectives, such as the impact or chances of pruning for interpretability and out-of-distribution generalization and detection capabilities \cite{zhu2023unleashing}. Such endeavours hold promise for advancing safety metrics in pruning practices.

We aim to raise awareness of cross-discipline safety challenges in model compression, uncertainty calibration, and robustness, serving as an initial step towards exploring common solutions.
\section{Acknowledgement} \label{sec:acknowledgement}
The research leading to these results is funded by the German Federal Ministry of Education and Research (BMBF) within the project CeCaS (\enquote{Central Car Server-Supercomputing}). The authors would like to thank the consortium for the successful cooperation.
{
    \small
    \bibliographystyle{ieeenat_fullname}
    \bibliography{main}

}


\end{document}